\documentclass[a4paper,fleqn]{cas-sc}

\usepackage[authoryear,longnamesfirst]{natbib}

\usepackage{times}
\usepackage{epsfig}
\usepackage{latexsym}
\usepackage{graphicx}
\usepackage{multirow}
\usepackage{multicol}
\usepackage{xcolor,colortbl}
\usepackage{booktabs}
\usepackage{amssymb}
\usepackage{subfigure}
\usepackage{hyperref}

\def\tsc#1{\csdef{#1}{\textsc{\lowercase{#1}}\xspace}}
\tsc{WGM}
\tsc{QE}
\tsc{EP}
\tsc{PMS}
\tsc{BEC}
\tsc{DE}

\begin{document}
\let\WriteBookmarks\relax
\def\floatpagepagefraction{1}
\def\textpagefraction{.001}
\shorttitle{}
\shortauthors{Yang et~al.}

\title [mode = title]{A Multi-task Learning Model for Chinese-oriented Aspect Polarity Classification and Aspect Term Extraction}



\author[1]{Heng Yang}[type=editor,
                        auid=000,bioid=1,
                        orcid=0000-0002-6831-196X]
\cormark[1]
\ead[url]{https://github.com/yangheng95/LCF-ATEPC}

\credit{Conceptualization, Methodology, Software, Draft preparation, Formal analysis}

\address[1]{School of Computer, South China Normal University, Guangzhou 510631, China}

\author[2]{Biqing Zeng}[orcid=0000-0001-9088-4759]

\credit{Methodology, Draft preparation, Resources}

\author[2]{JianHao Yang}[]
\credit{Data curation, Experiments}

\address[2]{School of Software, South China Normal University, Foshan 528225, China}

\author[3]{Youwei Song}[]
\credit{Experiments Advice, Paper Review}

\address[3]{Baidu Inc., Beijing 100085, China}

\author[1]{Ruyang Xu}[]
\credit{Investigation, Paper Review}

%

\cortext[cor1]{Biqing Zeng (email: \href{mailto:zengbiqing@scnu.edu.cn}{zengbiqing@scnu.edu.cn})}
\cortext[cor2]{Heng Yang (email: \href{mailto:yangheng@m.scnu.edu.cn}{yangheng@m.scnu.edu.cn})}
\fntext[fn1]{Email Address: \href{mailto:yangjianhao@m.scnu.edu.cn}{yangjianhao@m.scnu.edu.cn} (Jianhao Yang), \href{mailto:songyouwei@baidu.com}{songyouwei@baidu.com} (Youwei Song)  \href{mailto:cs\_xuruyang@m.scnu.edu.cn}{cs\_xuruyang@m.scnu.edu.cn} (Ruyang Xu)}


\begin{abstract}
Aspect-based sentiment analysis (ABSA) task is a multi-grained task of natural language processing and consists of two subtasks: aspect term extraction (ATE) and aspect polarity classification (APC). Most of the existing work focuses on the subtask of aspect term polarity inferring and ignores the significance of aspect term extraction. Besides, the existing researches do not pay attention to the research of the Chinese-oriented ABSA task. Based on the local context focus (LCF) mechanism, this paper firstly proposes a multi-task learning model for Chinese-oriented aspect-based sentiment analysis, namely LCF-ATEPC. Compared with existing models, this model equips the capability of extracting aspect term and inferring aspect term polarity synchronously, moreover, this model is effective to analyze both Chinese and English comments simultaneously and the experiment on a multilingual mixed dataset proves its availability. By integrating the domain-adapted BERT model, the LCF-ATEPC model achieved the state-of-the-art performance of aspect term extraction and aspect polarity classification in four Chinese review datasets. Besides, the experimental results on the most commonly used SemEval-2014 task4 Restaurant and Laptop datasets outperform the state-of-the-art performance on the ATE and APC subtask.
\end{abstract}
\label{abstract}

\begin{graphicalabstract}
\includegraphics{figs/LCF-ATEPC-new}
\end{graphicalabstract}

\begin{highlights}
	
\item Proposing a model for the joint task of aspect term extraction and aspect polarity classification.

\item The model proposed is Chinese language-oriented and applicable to the English language, with the ability to handle both Chinese and English reviews.

\item The model also integrates the domain-adapted BERT model for enhancement. 

\item The model achieves state-of-the-art performance on seven ABSA datasets.
\end{highlights}


\maketitle

\section{Introduction}
Aspect-based sentiment analysis \cite{pontiki2014semeval, pontiki2015semeval, pontiki2016semeval} (ABSA) is a fine-grained task compared with traditional sentiment analysis, which requires the model to be able to automatic extract the aspects and predict the polarities of all the aspects. For example, given a restaurant review: "The dessert at this restaurant is delicious but the service is poor," the full-designed model for ABSA needs to extract the aspects "dessert" and "service" and correctly reason about their polarity. 
In this review, the consumers' opinions on "dessert" and "service" are not consistent, with positive and negative sentiment polarity respectively. 

Generally, aspects and their polarity need to be manually labeled before running the aspect polarity classification procedure in the supervised deep learning models. However, most of the proposed models for aspect-based sentiment analysis tasks only focus on improving the classification accuracy of aspect polarity and ignore the research of aspect term extraction. Therefore, when conducting transfer learning on aspect-based sentiment analysis, those proposed models often fall into the dilemma of lacking aspect extraction method on targeted tasks because there is not enough research support. 

The APC task is a kind of classification problem. The researches concerning APC tasks is more abundant than the ATE task, and a large number of deep learning-based models have been proposed to solve APC problems, such as the models \cite{vo2015target, wagner2014dcu, tang2016effective, wang2016attention, ma2017interactive, fan2018multi} based on long short-term memory (LSTM) and the methodologies \cite{song2019attentional, zeng2019lcf} based on transformer \cite{vaswani2017attention}. The purpose of the APC task is to predict the exact sentiment polarity of different aspects in their context, rather than to fuzzily analyze the overall sentiment polarity on the sentence-level or document-level. In the APC task, the polarities are most usually classified into three categories: positive, negative, and neutral. It is obvious that the sentiment polarity classified based on aspects can better mine the fine-grained emotional tendency in reviews or tweets, thus providing a more accurate reference for decision-makers. 

Similar to the named entity recognition \cite{sang2003introduction} (NER) task, the ATE task is a sequence labeling task, which aims to extract aspects from the reviews or tweet. In most researches \cite{chen2017recurrent, xue2018aspect, chen2017improving}, the ATE task is studied independently, away from the APC task. The ATE task first segments a review into separate tokens and then infers whether the tokens belong to any aspect. The tokens may be labeled in different forms in different studies, but most of the studies have adopted the IOB\footnote{The labels adopted in this paper are: $B_{asp}, I_{asp}, O$} label to annotate tokens.

\begin{figure*}[pos=h]
	\centering
	\epsfig{file=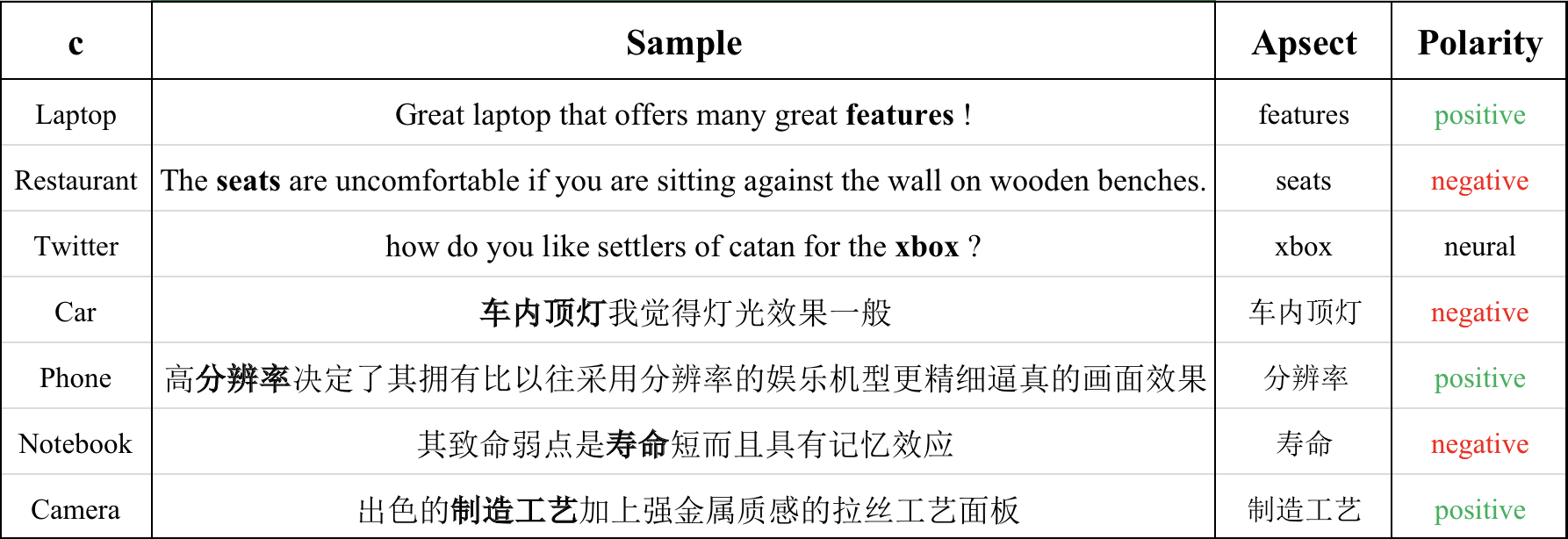, height=5cm}
	\caption{Several samples of the seven ATEPC datasets. All the datasets are domain-specific.}
	\label{fig:samples}
\end{figure*}

Aiming to automatically extract aspects from the text efficiently and analyze the sentiment polarity of aspects simultaneously, this paper proposes a multi-task learning model for aspect-based sentiment analysis. Multilingual processing is an important research orientation of natural language processing. The LCF-ATEPC\footnote{The codes for this paper are available at \url{https://github.com/yangheng95/LCF-ATEPC}} model proposed in this paper is a novel multilingual and multi-task-oriented model. Apart from achieving state-of-the-art performance in commonly used SemEval-2014 task4 datasets, the experimental results in four Chinese review datasets also validate that this model has a strong ability to expand and adapt to the needs of multilingual task. The proposed model is based on multi-head self-attention (MHSA) and integrates the pre-trained BERT \cite{devlin2019bert} and the local context focus mechanism, namely LCF-ATEPC. By training on a small amount of annotated data of aspect and their polarity, the model can be adapted to a large-scale dataset, automatically extracting the aspects and predicting the sentiment polarities. In this way, the model can discover the unknown aspects and avoids the tedious and huge cost of manually annotating all aspects and polarities. It is of great significance for the field-specific aspect-based sentiment analysis.

The main contributions of this article are as follows:

\begin{enumerate}
	
	\item For the first time, this paper studies the multi-task model of APC subtask and ATE subtask for multilingual reviews, which provides a new idea for the research of Chinese aspect extraction. 
	
	\item This paper firstly applies self-attention and local context focus techniques to aspect word extraction task, and fully explore their potential in aspect term extraction task.
	
	\item The LCF-ATEPC model proposed in this paper integrates the pre-trained BERT model, significantly improves both the performance of ATE task and APC subtask, and achieves new state-of-the-art performance especially the F1 score of ATE task. Besides, we adopted the domain-adapted BERT model trained on the domain-related corpus to the ABSA joint-task learning model. The experimental results show that the domain-adapted BERT model significantly promotes the performance of APC tasks on the three datasets, especially the Restaurant dataset.
	
	\item We designed and applied dual labels for the input sequence applicable for the SemEval-2014 and Chinese review datasets of ABSA joint-task, the aspect term label, and the sentiment polarity label, respectively. The dual label improves the learning efficiency of the proposed model.
	
\end{enumerate}

\section{Related Works}

Most ABSA-oriented methodologies regard the ATE and the APC as independent tasks and major in one of them. Accordingly, this section will introduce the related works of ATE and APC in two parts.

\subsection{Aspect Term Extraction}

The approaches to ATE tasks are classified into two categories: the early dictionary-based or rule-based approaches, and methodologies based on machine-learning or deep learning. 
\Citet{poria2014rule} proposed a new rule-based approach to extracting aspects from product reviews using common sense and sentence dependency trees to detect explicit and implicit aspects. 
\Citet{liu2015automated} adopts an unsupervised and domain-independent aspect extraction method that relies on syntactic dependency rules and can selects rules automatically.

Compared with manually annotating all aspects in the dataset, the models for ATE can learn the features of aspects and automatically extract aspects in the text, which greatly saves labor and time.
\Citet{mukherjee2012aspect} proposed a model that can extract and cluster aspects simultaneously according to the seed words provided by users for several aspect categories. By classification, synonymous aspects can be grouped into the same category.
\Citet{poria2016aspect} proposed the first aspect-oriented deep learning model in opinion mining, which deploys a 7-layer deep convolutional neural network to mark each word in the sentences with opinions as an aspect or non-aspect word.
\Citet{he2017unsupervised} proposed a new method for aspect term extraction, which utilizes word embedding to explore the co-occurrence distribution of words and applies the attention mechanism to weaken the irrelevant words and further improves the coherence of all aspects.
\Citet{wang2017coupled} proposed a deep neural network-based model namely coupled multilevel attention, which does not require any parser or other linguistic resources to be pre-processed and provides an end-to-end solution. Besides, the proposed model is a multi-layer attention network, where each layer deploys a pair of attentions. This model allows the aspect terms and opinion terms learned interactively and dual propagate during the training process.

For the Chinese-oriented ATE task, a multi-aspect bootstrapping (MAB) method \cite{zhu2011aspect} is proposed to extract the aspects of Chinese restaurant reviews. \Citet{zhao2015effect} introduced machine learning methods to explore and extract aspect terms from Chinese hotel reviews. they chose the optimal feature-dimension, feature representation, and maximum entropy (ME) classifier according to the empirical results, and studied the integral effect of aspect extraction.

Up to now, the MHSA and pre-trained model has not been applied in the ATE task. This paper explores the potential of the new techniques of deep learning and new network architecture in the ATE task.

\subsection{Aspect Polarity Classification}

Aspect polarity classification is another important subtask of ABSA. The approaches designed for the APC task can be categorized into traditional machine learning and recent deep learning methods.The APC task has been comprehensively turned to the the deep neural networks. Therefore, this section mainly introduces approaches based on deep learning techniques.

The most commonly applied deep neural network architectures for APC task are recurrent neural networks \cite{tang2016effective, wang2016attention, ma2017interactive, li2018transformation, huang2018aspect} (RNNs) and convolutional neural networks (CNNs) \cite{xue2018aspect, chen2017improving, zhang2018textual}. 
TD-LSTM \cite{tang2016effective} first divides the context of aspects into the left and right parts and modeling for them independently.
Attention mechanism \cite{bahdanau2014neural} has been adapted to APC task in the last few years. ATAE-LSTM takes the feature representation of aspects and context words as the input of the model and applies an attention mechanism to dynamically calculate the attention weight according to the relationship between aspects and context words, and finally predicts the polarity of aspects according to the weighted context features.
Another LSTM-based model IAN \cite{ma2017interactive} deployed with attention mechanism equips two independent LSTM networks to capture the features of the context and aspect, with interactively integrating and learning the inner correlation of the features of context and targeted aspects. The RAM \cite{chen2017recurrent} is a bi-directional LSTM-based architecture deploys a multi-layer deep neural network with dedicated memory layers. The multi-layer network utilizes the token features learned based on the attention mechanism and GRUs to finally obtain the global semantic features of the text to predict the sentiment polarities of targeted aspects. In order to retard the loss of context features during the training process, TNet \cite{li2018transformation} introduced a conventional transformation architecture based on context-preserving transformation (CPT) units. TNet integrates the bidirectional LSTM network and convolutional neural network and significantly improves the accuracy of sentiment polarity prediction. Multi-grained attention network \cite{fan2018multi} (MGAN) is a new deep neural network model, which equips with a variety of fine-grained attention mechanisms, and applies the fine-grained attention mechanisms to interactively learn the token-level features between aspects and context, making great use of the inherent semantic correlation of aspects and context.

\Citet{peng2018learning} proposed the methods for the Chinese language APC task, which conducted the APC task at the aspect level via three granularities. Two fusion methods for the granularities in the Chinese APC task are introduced and applied. Empirical results show that the proposed methods achieved promising performance on the most commonly used ABSA datasets and four Chinese review datasets. Meanwhile, a joint framework aimed to aspect sentiment classification subtask and aspect-opinion pair identification subtask is proposedby \Citet{chen2019Knowledge}, in which the external knowledge are considered and put into the network to alleviate the problem of insufficient train data. 
The gated alternate neural network (GANN) \cite{liu2019aspect} proposed for APC task aimed to solve the shortcomings of traditional RNNs and CNNs. The GANN applied the gate truncation RNN (GTR) to learn the aspect-dependent sentiment clue representations. 
\Citet{zeng2019joint} proposed an end-to-end neural network model for the ABSA task based on joint learning, and the experimental results on a Chinese review show that the proposed model works fine while conducting ATE and APC subtask simultaneously.

BERT-SPC is the BERT text pair classification model, it is a variation model of BERT and is adapted to solve the ABSA task in \cite{song2019attentional} and achieve high performance. LCF-BERT \cite{zeng2019lcf} proposed a feature-level local context focus mechanism based on self-attention, which can be applied to aspect level emotion analysis and many other fine-grained natural language processing tasks. BERT-ADA \cite{rietzler2019adapt} shows that although the pre-trained model based on a large universal corpus, and is easy to be applied to most tasks and improve performance. Still, it is not task-specific. For specific tasks, if the pre-trained BERT is adapted to specific tasks through the fine-tuning process on a task-related corpus, the task performance can be further improved.

\section{Methodology}

Aspect-based sentiment analysis relies on the targeted aspects, and most existing studies focus on the classification of aspect polarity, leaving the problem of aspect term extraction. To propose an effective aspect-based sentiment analysis model based on multi-task learning, we adopted domain-adapted BERT model from BERT-ADA and integrated the local context focus mechanism into the proposed model. This section introduces the architecture and methodology of LCF-ATEPC.

This section introduces the methodology of the APC module and the ATE module, respectively. and the contents are organized by order of the network layer hierarchy.

\subsection{Task Definition}
\subsubsection{Aspect Term Extraction}
Similar to name entity recognition (NER) task, the ATE task is a kind of sequence labeling task, and prepare the input based on IOB labels. We design the IOB labels as $B_{asp}, I_{asp}, O$, and the labels indicate the beginning, inside and outside of the aspect terms, respectively. For ATE task, the input of the example review ``The price is reasonable although the service is poor.'' will be prepared as $S=\{w_1,w_2 \cdots w_n\}$, and $w$ stands for a token after tokenization, $n=10$ is the total number of tokens. The example will be labeled in $Y=\{O, B_{asp}, O, O, O, O, B_{asp}, O, O, O\}$. 

\subsubsection{Aspect Polarity Classification}
Aspect polarity classification is a multi-grained sub-task of sentiment analysis, aiming at predicting the aspect polarity for targeted aspects. Suppose that ``The price is reasonable although the service is poor . '' is the input for APC task, consistently with ATE task, $S=\{w_1,w_2 \cdots w_n\}$ stands for all the token of the review, and $S^t=\{w_i,w_{i+1} \cdots w_{j}\} (1<=i<j<=n)$ is the aspect sequence within $S$, $i$ and $j$ are the beginning and end positions in $S$ respectively. 

\subsection{Model Architecture}

Aiming at the problem of insufficient research on aspect term extraction task, a joint deep learning model is designed in this section. This model combines aspect polarity classification task and aspect term extraction task, and two independent BERT layers are adopted to model the global context and the local context respectively. For conducting multi-task training at the same time, the input sequences are tokenized into different tokens and the each token is assigned two kinds of label. The first label indicates whether the token belongs to an aspect; the second label marks the polarity of the tokens belongs to the aspect.

\begin{figure}
	\centering
	\includegraphics[width=0.5\columnwidth]{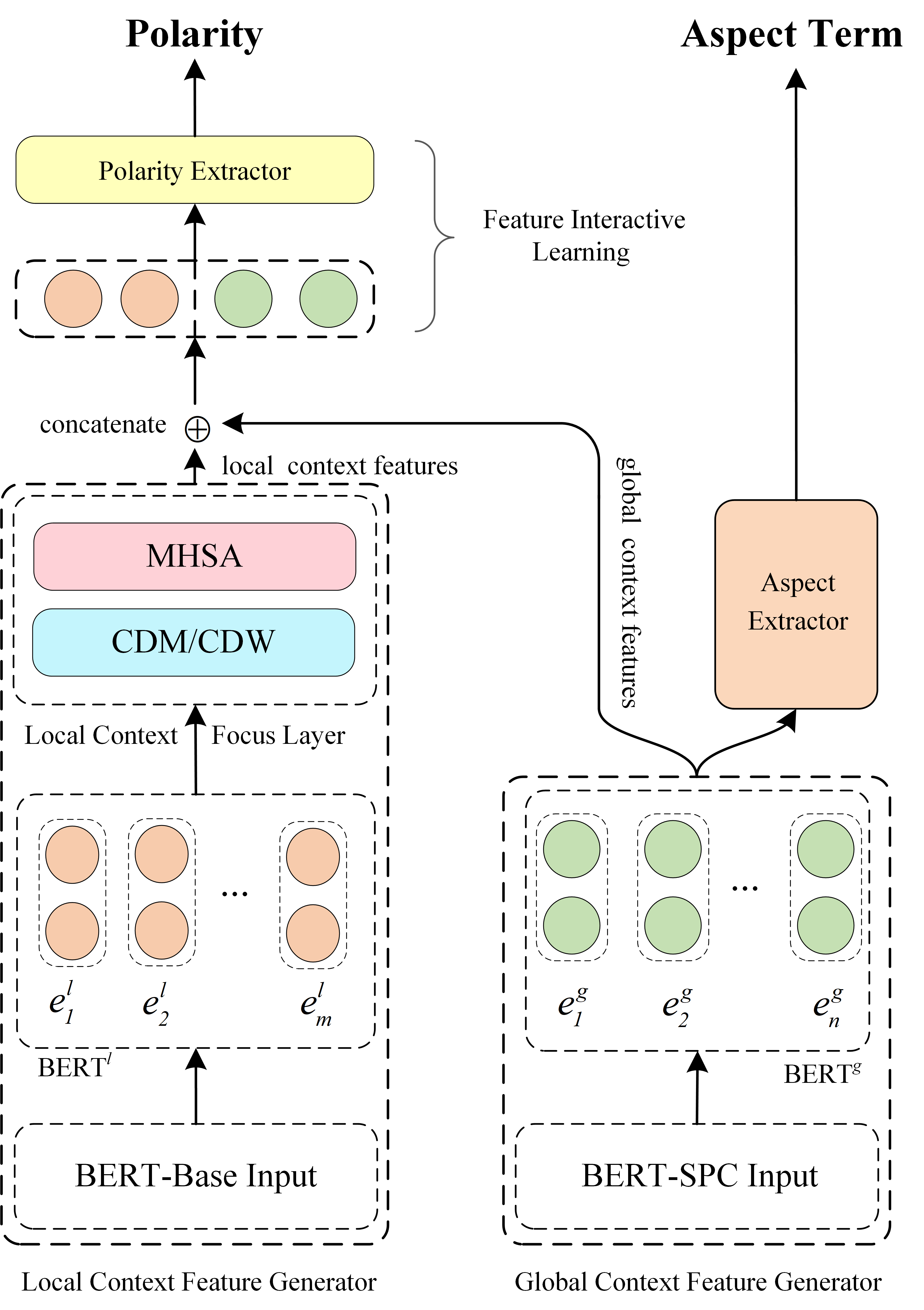}
	\caption{Network architecture of LCF-ATEPC}
	\label{fig:lcf-atepc}
\end{figure}

Fig \ref{fig:lcf-atepc} is the network architecture of LCF-ATEPC. 
Local context feature generator (LCFG) unit is on the left and a global context feature generator (GCFG) unit is on the right. Both context feature generator units contain an independent pre-trained BERT layer, $BERT^l$ and $BERT^g$ respectively. The LCFG unit extracts the features of the local context by a local context focus layer and a MHSA encoder. The GCFG unit deploys only one MHSA encoder to learn the global context feature. The feature interactive learning (FIL) layer combines the learning of the interaction between local context features and global context features and predicts the sentiment polarity of aspects. The extraction of aspects based on the features of the global context. 

\subsubsection{BERT-Shared Layer}

The pre-trained BERT model is designed to improve performance for most NLP tasks, and The LCF-ATEPC model deploys two independent BERT-Shared layers that are aimed to extract local and global context features. 
For pre-trained BERT, the fine-tuning learning process is indispensable. Both BERT-Shared layers are regarded as embedded layers, and the fine-tuning process is conducted independently according to the joint loss function of multi-task learning.
$X^{l}$ and $X^{g}$ are used to represent the tokenized inputs of LCFG and GCFG respectively, and we can obtain the preliminary outputs of local and global context features.
\begin{equation}
O^{l}_{BERT}=B E R T^{l}\left(X^{l}\right)
\end{equation}
\begin{equation}
O^{g}_{BERT}=B E R T^{g}\left(X^{g}\right)
\end{equation}

$O^{l}_{BERT}$ and $O^{g}_{BERT}$ are the output features of the LCFG and the GCFG, respectively. $BERT^{l}$ and $BERT^{g}$ are the corresponding BERT-shared layer embedded in the LCFG and the GCFG respectively.

\subsection{Multi-Head Self-Attention}

Multi-head self-attention is based on multiple scale-dot attention (SDA), which can be utilized to extract deep semantic features in the context, and the features are represented in self-attention score. The MHSA can avoids the negative influence caused by the long distance dependence of the context when learning the features.
Suppose $X_{SDA}$ is the input features learned by the LCFG. The scale-dot attention is calculate as follows:
\begin{equation}
\mathrm{SDA(X_{SDA})}=Softmax\left(\frac{Q \cdot K^{T}}{\sqrt{d_{k}}}\right) \cdot V 
\end{equation}
\begin{equation}
Q, K, V=f_{x}(X_{SDA})
\end{equation}
\begin{equation}
f_{x}(X_{SDA})=\left\{\begin{array}{l}{Q=X_{SDA} \cdot W^{q}} \\ {K=X_{SDA} \cdot W^{k}} \\ {V=X_{SDA} \cdot W^{v}}\end{array}\right.
\end{equation}

$Q$, $K$ and $V$ are the abstract matrices packed from the input features of SDA by three weight matrices $W_{q} \in \mathbb{R}^{d_{h} \times d_{q}}$, $W_{k} \in \mathbb{R}^{d_{h} \times d_{k}}$, $W_{v} \in \mathbb{R}^{d_{h} \times d_{v}}$. The MHSA performs multiple scaled-dot attention in parallel and concatenate the output features, then transform the features by multiplying a vector $W^{M H}$. $h$ represents the number of the attention heads and equal to 12. 
\begin{equation}
MHSA(X)=\tanh\left(\left\{H_{1};\ldots;H_{h}\right\}\cdot W^{MH}\right)
\end{equation}
The ``;'' means feature concatenation of each head. $W^{M H} \in \mathbb{R}^{hd_{v} \times d_{h}}$ is the parameter matrices for projection . Additionally, we apply a $\tanh$ activation function for the MHSA learning process, which significantly enhanced feature-capture capability.

\subsection{Local Context Focus}

\subsubsection{Semantic-Relative Distance}
\label{sec:SRD}

The determination of local context depends on semantic-relative distance (SRD), which is proposed to determine whether the context word belongs to the local context of a targeted aspect to help the model capture the local context. Local context is a new concept that can be adapted to most fine-grained NLP tasks. In the ABSA field, existing models generally segment input sequences into aspect sequences and context sequences, treat aspects and context as independent segments and model their characteristics separately. Instead of leaving the aspect alone as part of the input, this paper mines the aspect and its local context, because the empirical result shows the local context of the target aspect contains more important information.

SRD is a concept based on token-aspect pairs, describing how far a token is from the aspect. It counts the number of tokens between each specific token towards a targeted aspect as the SRD of all token-aspect pairs. The SRD is calculated as:
\begin{equation}
SRD_{i} = |i-P_{a}|-\lfloor\frac{m}{2}\rfloor
\end{equation}
where $i$ $(1<i<n)$ is the position of the specific token, $P_{a}$ is the central position of aspect. $m$ is the length of targeted aspect, and $SRD_{i}$ represents for the SRD between the $ i $-th token and the targeted aspect.

\begin{figure*}[pos=h]
	\centering
	\epsfig{file=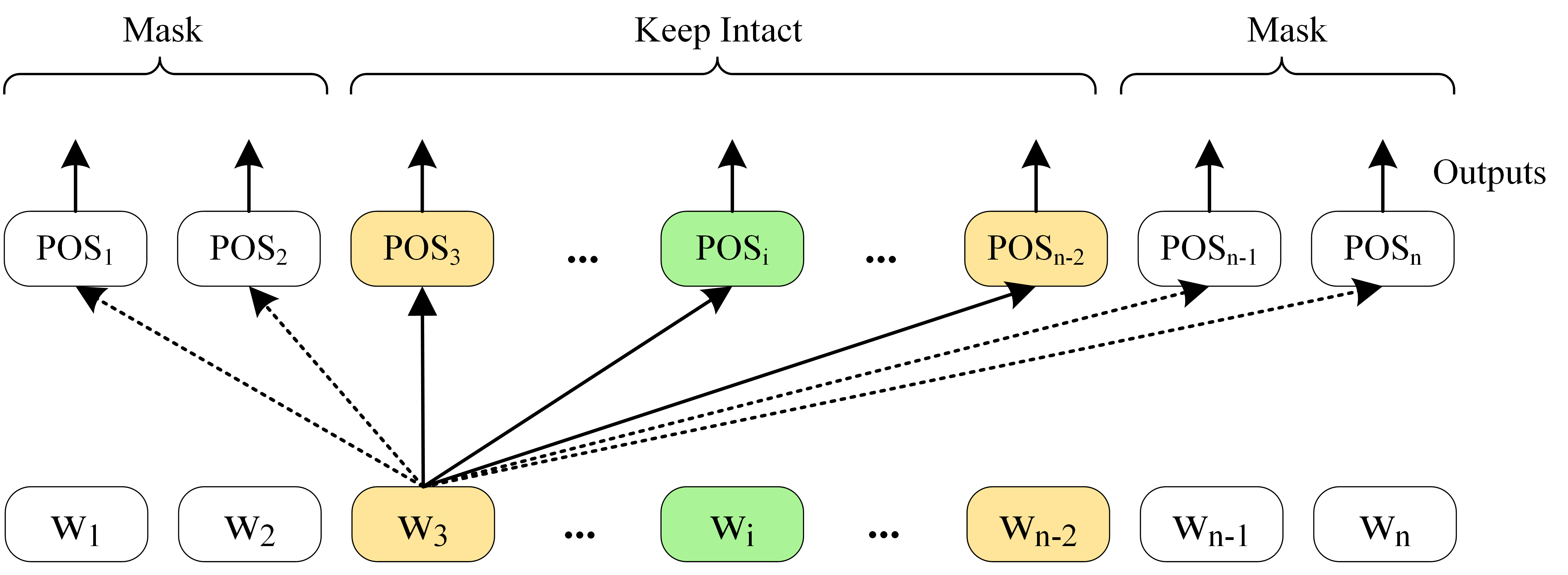,height=5cm}
	\caption{The simulation of the context-feature dynamic mask (CDM) mechanism. The arrows mean the contribution of the token in the computation of the self-attention score to arrowed positions (POS). And the features of the output position that the dotted arrow points to will be masked.}
	\label{fig:LCF1}
\end{figure*}

\begin{figure*}[pos=h]
	\centering
	\epsfig{file=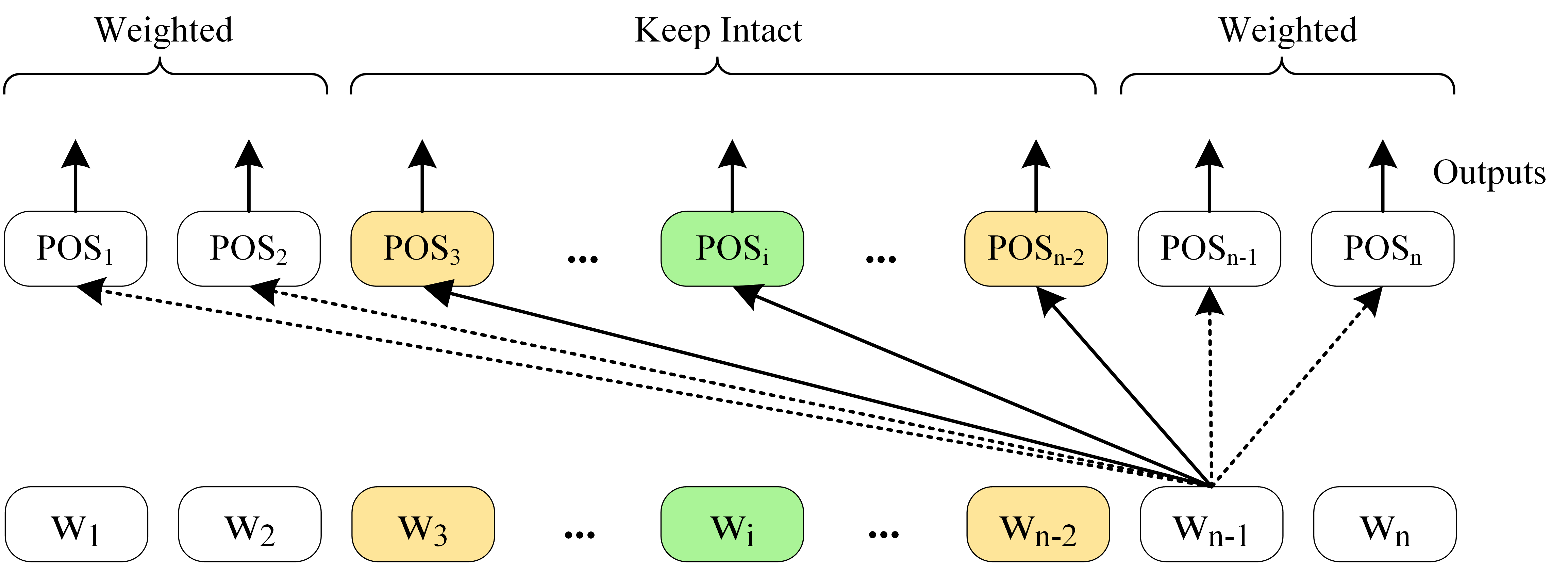,height=5cm}
	\caption{The simulation of the context-feature dynamic weighting (CDW) mechanism. The features of the output position (POS) that the dotted arrow points to will be weighted decay.}
	\label{fig:LCF2}
\end{figure*}

Figure \ref{fig:LCF1} and Figure \ref{fig:LCF2} are two implementations of the local context focus mechanism, the context-feature dynamic mask (CDM) layer and context-feature dynamic weighting (CDW) layer, respectively. The bottom and top of the figures represent the feature input and output positions (POS) corresponding to each token. The self-attention mechanism treats all tokens equally, so that each token can generate the self-attention score with other tokens through parallel matrix operation. According to the definition of MHSA, the features of the output position corresponding to each token are more closely related to itself. After calculating the output of all tokens by MHSA encoder, the output features of each output position will be masked or attenuated, except that the local context will be retained intact.

\subsubsection{Context-features Dynamic Mask}

Apart from to the features of the local context, the CDM layer will mask non-local context's features learned by the $BERT^l$ layer. Although it is easy to directly mask the non-local context words in the input sequence, it is inevitable to discard the features of non-local context words. As the CDM layer is deployed, only a relatively small amount of the semantic context itself will be masked at the corresponding output position. The relative representation of context words and aspects with relatively few semantics is preserved in the corresponding output position.

According to the CDM implementation, the features on all the positions of non-local context words will be set to zero vectors. In order to avoid the unbalanced distribution of features after the CDM operation, an MHSA encoder is utilized to learn and rebalance the masked local context features. Suppose that the $O_{BERT^l}$ is the preliminary output features of $BERT^l$, then we get the local context feature output as follows,
\begin{equation}
V_{i}=\left\{\begin{array}{ll}{E} & {SRD_{i} \leq \alpha} \\ {O} & {SRD_{i}>\alpha}\end{array}\right.
\end{equation}
\begin{equation}
M=\left[V_{1}^{m}, V_{2}^{m}, \ldots V_{n}^{m}\right]
\end{equation}
\begin{equation}
O_{C D M}^{l}=O_{BERT^l} \cdot M
\end{equation}

To mask the features of non-local context, we defines a feature masking matrix $M$, and $ V_{i}^{m} $ is the mask vectors for each token in the input sequence. $\alpha$ is the SRD threshold and $n$ is the length of input sequence including aspect. Tokens whose SRD regarding to the targeted aspect is less than the threshold $\alpha$ are the local contexts. The $E \in \mathbb{R}^{d_{h}}$ represents the ones vector and $O \in \mathbb{R}^{d_{h}}$ is the zeros vectors. ``$.$'' denotes the dot-product operation of the vectors.
\begin{equation}
O^{l}=M H S A\left(O_{C D M}^{l}\right)
\end{equation}

Finally the local context features learned by the CDM layer are delivered as $O^{l}$.

\subsubsection{Context-features Dynamic Weighting}

Although empirical results show that the CDM has achieved excellent performance compared with existing models, we design the CDW to explore the potential of LCF mechanism. The CDW is another implementation of the LCF mechanism, takes a more modest strategy compared to the CDM layer, which simply drops the features of the non-local context completely. 
While the features of local context retained intact, the features of the non-local context words will be weighted decay according to their SRD concerning a targeted aspect.

\begin{equation}
V_{i}=\left\{\begin{array}{cc}{E} & { S R D_{i} \leq \alpha} \\ {\frac{n -( SRD_{i}-\alpha)}{n} \cdot E} & {S R D_{i}>\alpha}\end{array}\right.
\end{equation}
\begin{equation}
W=\left[V_{1}^{w}, V_{2}^{w}, \ldots V_{n}^{w}\right]
\end{equation}
\begin{equation}
O_{C D W}^{l}=O_{BERT^l} \cdot W
\end{equation}

where $W$ is the constructed weight matrix and $V_{i}^{w}$ is the weight vector for each non-local context words. Consistently with CDM, $SRD_{i}$ is the SRD between the \textit{i}-th context token and a targeted aspect. $n$ is the length of the input sequence. $\alpha$ is the SRD threshold. ``$.$'' denotes the vector dot-product operation.
\begin{equation}
O^{l}=M H S A\left(O_{C D W}^{l}\right)
\end{equation}
$O_{C D W}^{l}$ is the output of CDW layer. The CDM and CDW layers are independent, which mean they are alternative. Both the output features of CDM and CDW layers are denoted as $O^{l}$. Besides, we tried to concatenate the learned features of CDM and CDW layers and take linear transformation as the features of local context.
\begin{equation}
O_{fusion}^l=[O_{C D M}^{l};O_{C D W}^{l}]
\end{equation}
\begin{equation}
O_{fusion}^{l} = W^{f} \cdot O^{f} + b^{f}
\end{equation}
\begin{equation}
O^{l}=M H S A\left(O_{fusion}^l\right)
\end{equation}

$W^{f}$, $O^{f}$ and $b^{f}$ are weight matrix and bias vector, respectively. The model can choose one of the three approaches to learn the local context features.
\subsection{Feature Interactive Learning}

LCF-ATEPC does not only rely on local context features for sentiment polarity classification, but combines and learns the local context features and the global context features to conduct polarity classification.

\begin{equation}
O^{l g}=\left[O^{l} ; O^{g}\right]
\end{equation}
\begin{equation}
O_{dense}^{l g} = W^{lg} \cdot O^{lg} + b^{lg}
\label{dense}
\end{equation}
\begin{equation}
O_{FIL}^{l g}=M H S A\left(O_{dense}^{lg}\right)
\end{equation}

$O^{l} $ and $ O^{g}$ are the local context features and global context features, respectively. $ W^{lg} \in \mathbb{R}^{d_{h} \times 2d_{h}}$ and $ b^{lg} \in \mathbb{R}^{d_{h}}$ are the weights and bias vectors, respectively. To learn the features of the concatenated vectors, an MHSA encoding process is performed on the $O_{dense}^{l g}$. 

\subsection{Aspect Polarity Classifier}
Aspect polarity classifier performs a head-pooling on the learned concatenated context features. Head-pooling is to extract the hidden states on the corresponding position of the first token in the input sequence. then a Softmax operation is applied to predict the sentiment polarity.
\begin{equation}
X_{pool}^{lg}=POOL\left(O_{FIL}^{l g}\right)
\end{equation}
\begin{equation}
Y_{polarity}=\frac{\exp (X_{pool}^{l g})} {\sum_{k=1}^{C}\exp(X_{pool}^{lg})}
\end{equation}
where $C$ is the number of sentiment categories, and $Y_{polarity}$ represents the polarity predicted by aspect polarity classifier.

\subsection{Aspect Term Extractor}
Aspect term extractor first performs the token-level classification for each token, suppose $T_{i}$ is the features on the corresponding position of token $T$,
\begin{equation}
Y_{term}=\frac{\exp (T_{i})} {\sum_{k=1}^{N}\exp(T_{i})}
\end{equation}
where $N$ is the number of token categories, and $Y_{term}$ represents the token category inferred by aspect polarity classifier.

\subsection{Training Details}

The LCFG and the GCFG are based on the BERT-BASE models. However, the BERT-SPC \cite{song2019attentional} model significantly improved the APC tasks and can be adapted to enhance the APC performance of LCF-ATEPC model. The BERT-SPC only refactored the input sequence form compared with BERT-BASE model. The input sequence of BERT-BASE is formed in \textit{``[CLS]'' + sequence + ``[SEP]''}, while it is formed in \textit{``[CLS]'' + sequence + ``[SEP]'' + aspect + ``[SEP]''} for BERT-SPC.

\begin{figure*}[pos=th]
	\centering
	
	\subfigure[Sample 1]
	{
		\epsfig{file=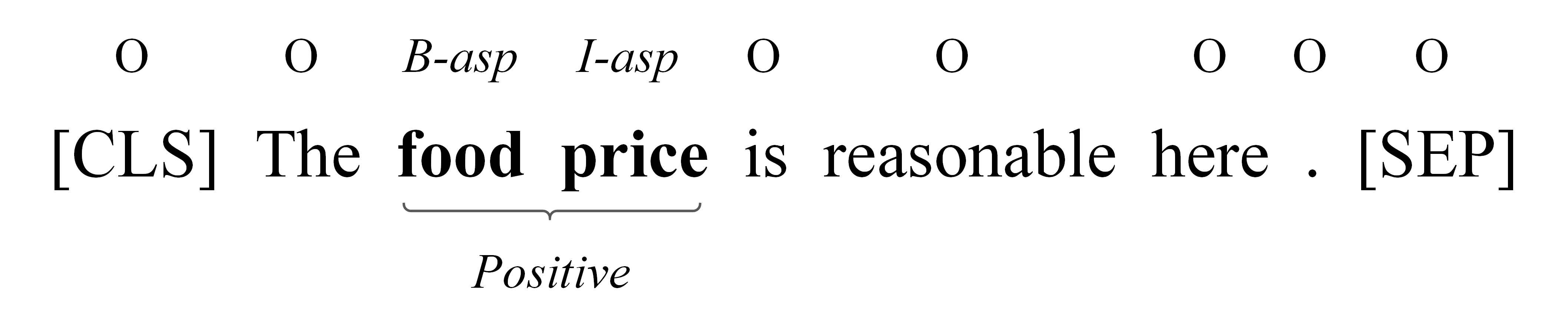,height=2cm}
		\label{fig:input}
	}
	\subfigure[Sample 2]
	{
		\epsfig{file=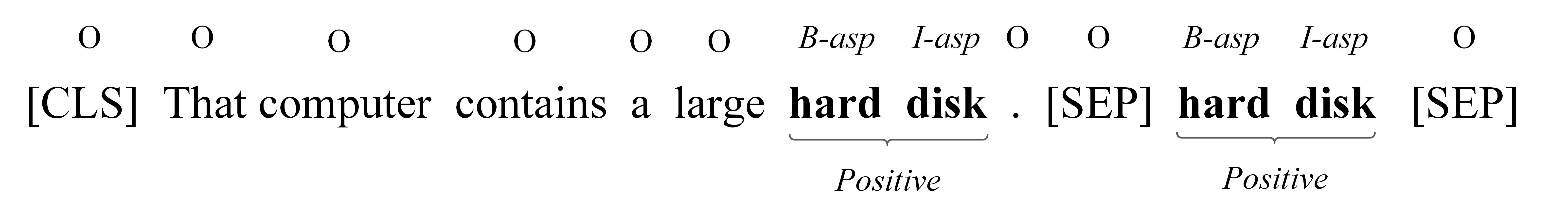,height=2cm}
		\label{fig:input1}
	}
	\subfigure[Sample 3]
	{
		\epsfig{file=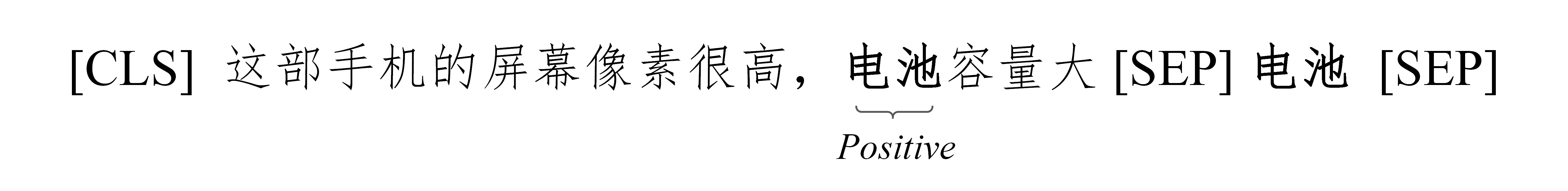,height=1.5cm}
		\label{fig:input2} 
	}
	\caption{ \ref{fig:input} is he input sequence formatted for the BERT-BASE model. The first line is composed of the aspect term labels; The second and third lines are the input sequence after tokenization and polarity label, respectively. \ref{fig:input1} represent the sample input sequence formatted for BERT-SPC model and the aspects in two position are labeled simultaneously. \ref{fig:input2} depicts a sample of Chinese sequence input format for BERT-SPC model.}
	\label{fig:inputs}
\end{figure*}

Since LCF-ATEPC is a multi-task learning model, we redesigned the form of data input and adopted dual labels of sentiment polarity and token category. The Figure \ref{fig:inputs} are the input samples of BERT-BASE and BERT-SPC model, respectively. 

The cross-entropy loss is adopted for APC and ATE subtask and the $\mathbf{L}_{2}$ regularization is applied in LCF-ATEPC, here is the loss function for APC task,
\begin{equation}
\mathcal{L}_{apc}=\sum_{1}^{C} \widehat{y_{i}} \log y_{i}+\lambda \sum_{\theta \in \Theta} \theta^{2}
\end{equation}
where $C$ is the number of polarity categories, $\lambda$ is the $L_{2}$ regularization parameter, and $\Theta$ is the parameter-set of the LCF-ATEPC. The loss function for ATE task is 
\begin{equation}
\mathcal{L}_{ate}=\sum_{1}^{N}\sum_{1}^{k}\widehat{t_{i}} \log t_{i}+\lambda \sum_{\theta \in \Theta} \theta^{2}
\end{equation}
where $N$ is the number of token classes and $k$ is the sum of the tokens in each input sequence. Accordingly, the loss function of LCF-ATEPC is as follows:

\begin{equation}
\mathcal{L}_{atepc} = \mathcal{L}_{apc}+ \mathcal{L}_{ate}
\end{equation}

\section{Experiments}

\subsection{Datasets and Hyperparameters Setting}

To comprehensive evaluate the performance of the proposed model, the experiments were conducted in three most commonly used ABSA datasets, the Laptops and Restaurant datasets of SemEval-2014 Task4 subtask2 \cite{pontiki2014semeval} and an ACL Twitter social dataset \cite{dong2014adaptive}. To evaluate our model capability with processing the Chinese language, we also tested the performance of LCF-ATEPC on four Chinese comment datasets \cite{che2015sentence, zhao2015principal, peng2018learning} (Car, Phone, Notebook, Camera). We preprocessed the seven datasets. We reformatted the origin dataset and annotated each sample with the IOB labels for ATE task and polarity labels for APC tasks, respectively. The polarity of each aspect on the Laptops, Restaurants and datasets may be positive, neutral, and negative, and the conflicting labels of polarity are not considered. The reviews in the four Chinese datasets have been purged, with each aspect may be positive or negative binary polarity. To verify the effectiveness and performance of LCF-ATEPC models on multilingual datasets, we built a multilingual dataset by mixing the 7 datasets. We adopt this dataset to conduct multilingual-oriented ATE and APC experiments.

The table demonstrates the details of these datasets\footnote{The dataset processed for joint ATE and APC task are available at \url{https://github.com/yangheng95/LCF-ATEPC}}. 

\begin{table}[pos=h]
	\small
	\centering
	\caption{The ABSA datasets for ATE and APC subtasks, including three English datasets and four Chinese datasets.}
		\begin{tabular}{ccccccc}
			\toprule
			\multirow{2}{*}{\textbf{Datasets}}&
			\multicolumn{2}{c}{\textbf{Positive}}&\multicolumn{2}{c}{\textbf{Negative}}&\multicolumn{2}{c}{\textbf{Neural}}\cr
			\cmidrule(lr){2-7}
			&\textbf{Train}&\textbf{Test}&\textbf{Train}&\textbf{Test}&\textbf{Train}&\textbf{Test} \cr
			\midrule
			Laptop      &994   &341    &870    &128    &464    &169    \cr
			Restaurant  &2164  &728    &807    &196    &637    &196    \cr
			Twitter     &1561  &173    &1560   &173    &3127   &346    \cr
			Car     &708  &164   &213   &66   &-   &-    \cr
			Phone     &1319  &341    &668   &156    &-   &-    \cr
			Notebook     &328  &88    &168   &35    &-   &-    \cr
			Camera     &1197  &322    &546   &113    &-   &-    \cr
			Multilingual Mixed &8271  &2157    &4340   &867    &4228   &711    \cr
			\bottomrule
		\end{tabular}
	\label{tab:datasets}
\end{table}

The samples distribution of those datasets is not balanced. For example, most samples in the restaurant dataset are positive, while the neutral samples in the Twitter dataset account for the majority.

\begin{table}[pos=h]
	\small
	\centering
	\caption{Global hyperparameters settings for the LCF-ATEPC model, BERT-BASE and BERT-SPC models in the experiments.}
	
	\begin{tabular}{cc}
		\toprule
		\textbf{Hyperparameters}&\textbf{Setting} \cr
		\midrule
		learning rate	&$3\times10^{-5}$       \cr
		batch size		&16  				    \cr
		training epochs	&5  			        \cr
		max sequence length	&80			        \cr
		SRD			&5 	        \cr
		optimizer   &AdamW  			        \cr
		\bottomrule
	\end{tabular}
	\label{tab:hyperparameters}
\end{table}

Apart from some hyperparameters setting referred to previous researches, we also conducted the controlled trials and analyzed the experimental results to optimize the hyperparameters setting. The superior hyperparameters are listed in Table \ref{tab:hyperparameters}. The default SRD setting for all experiments is 5, with additional instructions for experiments with different SRD.

\subsection{Compared Methods}


We compare the LCF-ATEPC model to current state-of-the-art methods. Experimental results show that the proposed model achieves state-of-the-art performance both in the ATE and APC tasks. \\
\textbf{ATAE-LSTM} \cite{wang2016attention} is a classical LSTM-based network for the APC task, which applies the attention mechanism to focus on the important words in the context. Besides, ATAE-LSTM appends aspect embedding and the learned features to make full use of the aspect features. The ATAE-LSTM can be adapted to the Chinese review datasets.\\
\textbf{ATSM-S} \cite{peng2018learning} is a baseline model of the ATSM variations for Chinese language-oriented ABSA task. This model learns the sentence and aspect terms at three perspectives of granularity. \\
\textbf{GANN} is novel neural network model for APC task aimed to solve the shortcomings of traditional RNNs and CNNs. The GANN applied the Gate Truncation RNN (GTR) to learn informative aspect-dependent sentiment clue representations. GANN obtained the state-of-the-art APC performance on the Chinese review datasets. \\
\textbf{AEN-BERT} \cite{song2019attentional} is an attentional encoder network based on the pretrained BERT model, which aims to solve the aspect polarity classification. \\
\textbf{BERT-PT} \cite{xu2019bert} is a BERT-adapted model for Review Reading Comprehension (RRC) task, a task inspired by machine reading comprehension (MRC), it could be adapted to aspect-level sentiment classification task. \\
\textbf{BERT-BASE} \cite{devlin2019bert} is the basic pretrained BERT model. We adapt it to ABSA multi-task learning, which equips the same ability to automatically extract aspect terms and classify aspects polarity as LCF-ATEPC model. \\
\textbf{BERT-SPC} \cite{song2019attentional} is a pretrained BERT model designed for the sentence-pair classification task and improves the APC subtask of LCF-ATEPC model. \\
\textbf{BERT-ADA} \cite{rietzler2019adapt} is a domain-adapted BERT-based model proposed for the APC task, which fine-tuned the BERT-BASE model on task-related corpus. This model obtained state-of-the-art accuracy on the Laptops dataset.\\
\textbf{LCF-ATEPC\footnote{We implement our model based on pytorch-transformers: \url{ https://github.com/huggingface/pytorch-transformers}}} is the multi-task learning model for the ATE and APC tasks, which is based on the the BERT-SPC model and local context focus mechanism. \\
\textbf{LCF-ATE} are the variations of the LCF-ATEPC model which only optimize for the ATE task. \\
\textbf{LCF-APC} are the variations of LCF-ATEPC and it only optimize for the APC task during training process.\\

\subsection{Results Analysis}
The experiments are conducted in several segments. First, the baseline performance of LCF-ATEPC on all Chinese and English data sets was tested, and then the effectiveness of multi-task learning was demonstrated. Finally, the assistance of domain-adapted BERT model in improving performance was evaluated and the sensitivity of different datasets to SRD was studied.

\subsubsection{Performance on Chinese Review Datasets}

Table \ref{tab:chineseresult} are the experimental results of LCF-ATEPC models on four Chinese review datasets.

\begin{table*}[pos=h]
	\centering
	\setlength{\tabcolsep}{1mm}{
		\caption{The experimental results (\%)  of LCF-ATEPC models on four Chinese datasets. ``-'' represents the result is not unreported. The optimal performance is in \textbf{bold}.}
		\begin{tabular}{ccccccccccccc}
			\toprule
			\multirow{2}[0]{*}{\textbf{Model}} & 
			\multicolumn{3}{c}{\textbf{Car}} & 
			\multicolumn{3}{c}{\textbf{Phone}} & \multicolumn{3}{c}{\textbf{Notebook}}&
			\multicolumn{3}{c}{\textbf{Camera}} \cr 
			\cmidrule(lr){2-4} 
			\cmidrule(lr){5-7} 
			\cmidrule(lr){8-10}
			\cmidrule(lr){11-13}
			& $F1_{ate}$ & $Acc_{apc}$ & $F1_{apc}$
			& $F1_{ate}$ & $Acc_{apc}$ & $F1_{apc}$ 
			& $F1_{ate}$ & $Acc_{apc}$ & $F1_{apc}$ 
			& $F1_{ate}$ & $Acc_{apc}$ & $F1_{apc}$\cr 
			\midrule
			ATAE-LSTM &- &81.90	&76.88	   &- &85.77	&83.87 &-  &83.47	&82.14	 &-	&85.54	&84.09	  \cr 
			ATSM-S &-   &82.94	&64.18   &- &84.86	&75.35 &-  &75.59	&60.09 &-	&82.88	&72.50  \cr 
			GANN &-  &83.71	&77.66   &- &89.17	&88.16 &- &82.65	&82.16 &-	&87.99	&86.75  \cr

			BERT-BASE   &86.9	&\textbf{98.26}	&97.84  &92.1	&97.18	&96.73 &84.62	&\textbf{94.31}	&\textbf{93.38}  &86.13	&97.47	&96.72  \cr 
			\midrule
			LCF-ATEPC-CDM  &\textbf{88.0}	&\textbf{98.26}	&\textbf{97.89 }&90.82	&97.18	&96.73  &88.89	&93.5	&92.38 &87.88	&\textbf{98.16}	&\textbf{97.61} \cr 
			LCF-ATEPC-CDW  &87.63	&\textbf{98.26}	&97.86  &91.19	&\textbf{97.38}	&\textbf{96.96 }&88.54	&\textbf{94.31}	&93.29  	&\textbf{88.47}	&97.47	&96.76\cr 
			LCF-ATEPC-Fusion  &86.64	&97.39	&96.72  &\textbf{92.55}	&\textbf{97.38}	&\textbf{96.96}  &\textbf{89.16}	&\textbf{94.31}	&93.29	&87.9	&96.78	&95.86 \cr 
			\bottomrule
		\end{tabular}
		\label{tab:chineseresult}
	}
\end{table*}

\subsubsection{Performance on SemEval-2014 task4}

Table \ref{tab:mainresult} lists the main experimental results of LCF-ATEPC models to compare the performance with other ABSA-oriented models.

\begin{table*}[pos=h]
	\centering
	\caption{Experimental results (\%) of the LCF-ATEPC model. $F1_{ate}$, $Acc_{apc}$ and $F1_{apc}$ are the macro-F1 score of ATE subtask, accuracy and macro-F1 score of the APC subtask. The unreported experimental results are denoted by ``-''. The  ``$\dagger$'' means the F1 score of the ATE task is not available for BERT-SPC input format and the optimal performance is in \textbf{bold}.}
	\setlength{\tabcolsep}{1mm}{
		\begin{tabular}{cccccccccccccccc}
			\toprule
			\multirow{2}[0]{*}{\textbf{Model}} & \multicolumn{3}{c}{\textbf{Laptop}} & \multicolumn{3}{c}{\textbf{Restaurant}}&\multicolumn{3}{c}{\textbf{Twitter}}
			&\multicolumn{3}{c}{\textbf{Multilingual Mixed}}
			 \cr 
			\cmidrule(lr){2-4} \cmidrule(lr){5-7} \cmidrule(lr){8-10} \cmidrule(lr){11-13}
			& $F1_{ate}$ & $Acc_{apc}$ & $F1_{apc}$ & $F1_{ate}$ & $Acc_{apc}$ & $F1_{apc}$ & $F1_{ate}$ & $Acc_{apc}$ & $F1_{apc}$ & $F1_{ate}$ & $Acc_{apc}$ & $F1_{apc}$ \cr 
			\midrule
			BERT-PT & -  &78.07 &75.08 & -  &84.95 &76.96 & - & - & - & - & - & -\cr 
			AEN-BERT & -  & 79.93 & 76.31 & -  & 83.12 & 73.76 & - & 74.71 & 73.13 & - & - & - \cr 
			SDGCN-BERT & - & 81.35 & 78.34 & -  & 83.57 & 76.47 & - & - & - & - & - & -\cr 
			BERT-BASE & 82.57  &79.4	&75.24   &88.6	&82.66	&74.13  &95.45	&76.27	&74.91 &85.35  &81.08	&76.63 \cr 
			BERT-SPC &$\dagger$  &79.56	&75.55    &$\dagger$ &\textbf{86.77}	&80.52 &$\dagger$  &76.27	&75.16  &$\dagger$ &81.08	&77.04 \cr 

			\midrule
			LCF-ATEPC-CDM &82.06   &80.03	&76.6  &88.49   &86.06	&80.22  &95.98   &75.11	&74.06  &85.24 &\textbf{82.39}	&\textbf{78.27} \cr 
			LCF-ATEPC-CDW &81.61   &80.03	&77.17  &88.65  &85.97	&80.42 &96.12  &76.56	&\textbf{75.04} &85.35 &81.43	&77.31 \cr 
			LCF-ATEPC-Fusion &\textbf{83.82} &\textbf{80.97}	&\textbf{77.86}  &\textbf{89.02} &\textbf{86.77}	&\textbf{80.54}  &\textbf{96.4} &\textbf{76.7}	&74.54  &\textbf{85.4 } &81.77	&77.38 \cr 
			\bottomrule
		\end{tabular} 
	}
	\label{tab:mainresult}
	
\end{table*}

The LCF-ATEPC models are multilingual-oriented. To demonstrate its ability to simultaneously input and analyze reviews in multiple languages, we constructed and experimented with a multilingual dataset fore-mentioned.
And result on the multilingual mixed dataset illustrates the effectiveness of the LCF-ATEPC models.

\subsection{Overall Performance Analysis}

Many models for ABSA tasks do not take into account the ATE subtask, but there are still some joint models \cite{nguyen2018joint} based on the traditional neural network architecture to conduct the APC and ATE tasks simultaneously. Benefit from the joint training process, the two ABSA subtasks of APC and ATE can promote each other and improve the performance.

The ATEPC-Fusion mechanism works better on most datasets, especially English review datasets.
Surprisingly, for the Laptop and Restaurant datasets, guests occasionally have a unified ``global'' view in a specific review. That is, if the customer is not satisfied with one aspect, it is likely to criticize the other. Things will be the same if a customer prefers a restaurant he would be tolerant of some small disamenity, so the ATEPC-CDW mechanism performs better because it does not completely mask the local context of the other aspect. In the multi-task learning process, the convergence rate of APC and ATE tasks is different, so the model does not achieve the optimal effect at the same time. 

We build a joint model for the multi-task of ATE and APC based on the BERT-BASE model. After optimizing the model parameters according to the empirical result, the joint model based on BERT-BASE achieved hopeful performance on all three datasets and even surpassed other proposed BERT based improved models on some datasets, such as BERT-PT, AEN-BERT, SDGCN-BERT, and so on. 
Compared with the BERT-BASE model, BERT-SPC significantly improves the accuracy and F1 score of aspect polarity classification. In addition, for the first time,LCF-ATEPC has increased the F1 score of ATE subtask on three English datasets up to 82\%, 89\%, 96\%, respectively.

ATEPC-Fusion is a supplementary scheme of LCF mechanism, and it adopts a moderate approach to generate local context features. The experimental results show that its performance is also better than the existing BERT-based models.

\subsubsection{Effectiveness of Multi-task Learning}

Keeping the main architecture of the LCF-ATEPC model unchanged, we tried to only optimize parameters for a single task in the multi-task model to explore the difference between the optimal performance of a single task and the multi-task learning model \footnote{The loss function of the LCF-ATEPC is set to $\mathcal{L}_{apc}$ and $\mathcal{L}_{ate}$ while optimizing for APC and ATE task, respectively. }. 

\begin{table*}[pos=th]
	\centering

		\caption{The empirical performance comparison between multi-task and single-task learning. The ``-'' indicates that the statistics are not important during single-task learning optimization and not listed in the table. The optimal performance is in \textbf{bold}.}
		\begin{tabular}{cccccccccc}
			\toprule
			\multirow{2}[0]{*}{\textbf{Model}} & \multicolumn{3}{c}{\textbf{Laptop}} & \multicolumn{3}{c}{\textbf{Restaurant}} & \multicolumn{3}{c}{\textbf{Twitter}} \cr 
			\cmidrule(lr){2-4} \cmidrule(lr){5-7} \cmidrule(lr){8-10}
			& $F1_{ate}$ & $Acc_{apc}$ & $F1_{apc}$ & $F1_{ate}$ & $Acc_{apc}$ & $F1_{apc}$ & $F1_{ate}$ & $Acc_{apc}$ & $F1_{apc}$ \cr 
			\midrule
			LCF-ATEPC-CDM &82.06   &80.03	&76.6   &88.49   &86.06	&80.22  &95.98   &75.11	&74.06  \cr 
			LCF-ATEPC-CDW &81.61   &80.03	&77.17  &88.65  &85.97	&80.42 &96.12  &76.56	&\textbf{75.04} \cr 
			LCF-ATEPC-Fusion &83.82 &\textbf{80.97}	&\textbf{77.86}  &89.02 &\textbf{86.77}	&80.54  &96.4 &\textbf{76.7}	&74.54\cr 
			\midrule
			LCF-ATE-CDM &\textbf{84.64 } &$\dagger$ &$\dagger$  &\textbf{89.53}  &$\dagger$ &$\dagger$   &\textbf{97.77} &$\dagger$  &$\dagger$ \cr
			LCF-APC-CDM &$\dagger$ &80.19	&76.52   &$\dagger$ &85.7	&79.99   &$\dagger$ &75.4	&74.43 \cr 
			LCF-APC-CDW &$\dagger$ &80.35	&76.23  &$\dagger$ &85.79	&79.17   &$\dagger$ &75.98	&74.49 \cr    
			LCF-APC-Fusion &$\dagger$ &80.5 &77.77 &$\dagger$ &86.15	&\textbf{80.76} &$\dagger$  &75.54	&74.55  \cr 
			\bottomrule
		\end{tabular}
		\label{tab:singletaskresult}
	
\end{table*}

The Figure \ref{tab:singletaskresult} depicts the performance of the LCF-ATEPC model when performing an single APC or ATE task. Experimental results show that on some datasets the LCF-ATEPC model performs better concerning APC or ATE single task than conducting ABSA multi-task on some datasets. In general, the proposed model LCF-ATEPC proposed in this paper is still superior to other ABSA-oriented multi-task models and even the single-task models aim to APC or ATE. When optimizing the model parameters for through back-propagation of multiple tasks, the multi-task learning model needs to take into account multiple loss functions of the different subtasks. So sometimes the multi-task learning cannot achieve as the best effect as single-task learning does, which is also the compromise of the multi-task learning model when dealing with multiple tasks.

\subsubsection{Domain-adaption for LCF-ATEPC}

\begin{table*}[pos=h]
	\centering

		\caption{Experimental results of LCF-ATEPC on Laptop and Restaurant data sets using domain-adapted pretrained BERT.}
		\begin{tabular}{ccccccc}
			\toprule
			\multirow{2}[0]{*}{\textbf{Model}} & \multicolumn{3}{c}{\textbf{Laptop}} & \multicolumn{3}{c}{\textbf{Restaurant}} \cr 
			\cmidrule(lr){2-4} \cmidrule(lr){5-7} 
			& $F1_{ate}$ & $Acc_{apc}$ & $F1_{apc}$ & $F1_{ate}$ & $Acc_{apc}$ & $F1_{apc}$  \cr 
			\midrule
			BERT-BASE &84.64  &79.72	&75.18 &89.76	&85.97	&77.68 \cr
			BERT-SPC &$\dagger$ &82.08	&78.34  &$\dagger$  &88.83	&83.52    \cr 
			BERT-ADA &-  &80.23  &75.77  &-  &87.14 &80.09    \cr 
			\midrule
			LCF-ATEPC-CDM &\textbf{85.29} &\textbf{83.02}	&\textbf{79.84}  &89.78  &\textbf{90.18} &\textbf{85.88}    \cr
			LCF-ATEPC-CDW &85.24  &81.76  &78.06  &\textbf{89.99}  &88.65	&83.7    \cr 
			LCF-ATEPC-Fusion &85.06  &81.76	&78.65 &89.94 &89.45	&84.76  \cr
			\bottomrule
		\end{tabular}
		\label{tab:domainadaptedresult}
	
\end{table*}

The BERT-BASE model is trained on a large-scale general corpus, so the fine-tuning during process during training process is significant and inevitable for BERT-based models. Meanwhile, the ABSA datasets commonly benchmarked are generally small with the domain-specific characteristic, the effect of BERT-BASE model on the most ABSA datasets can be further improved through domain-adaption.
Domain adaption is a effective technique while integrating the pre-trained BERT-BASE model. By further training the BERT-BASE model in a domain-related corpus similar to or homologous to the target ABSA dataset, then domain-related pretrained BERT model can be obtained. We adopted the method proposed in \cite{rietzler2019adapt} to obtain the domain-adapted pre-trained BERT model based on the corpus of Yelp Dataset Challenge reviews\footnote{This corpus are available at \url{https://www.yelp.com/dataset/challenge}} and the amazon Laptops review dataset\cite{he2016ups}.
Table \ref{tab:domainadaptedresult} shows that the performance of APC task significantly improved by domain-adapted BERT model. The accuracy benchmark in the classical Restaurant achieving more than 90\%, which means that the LCF-ATEPC is the first ABSA-oriented model obtained up to 90\% accuracy on the Restaurant dataset. In addition, experimental result on the Laptop dataset also prove the effectiveness of domain-adaption in multi-task learning.
Besides, the experimental results on the laptop dataset also validate the effectiveness of domain-adapted BERT model for ABSA multi-task learning.

\subsubsection{SRD Sensitivity on Different Datasets}

\begin{figure*}[pos=th]
	\centering
	\subfigure[]
	{
		\epsfig{file=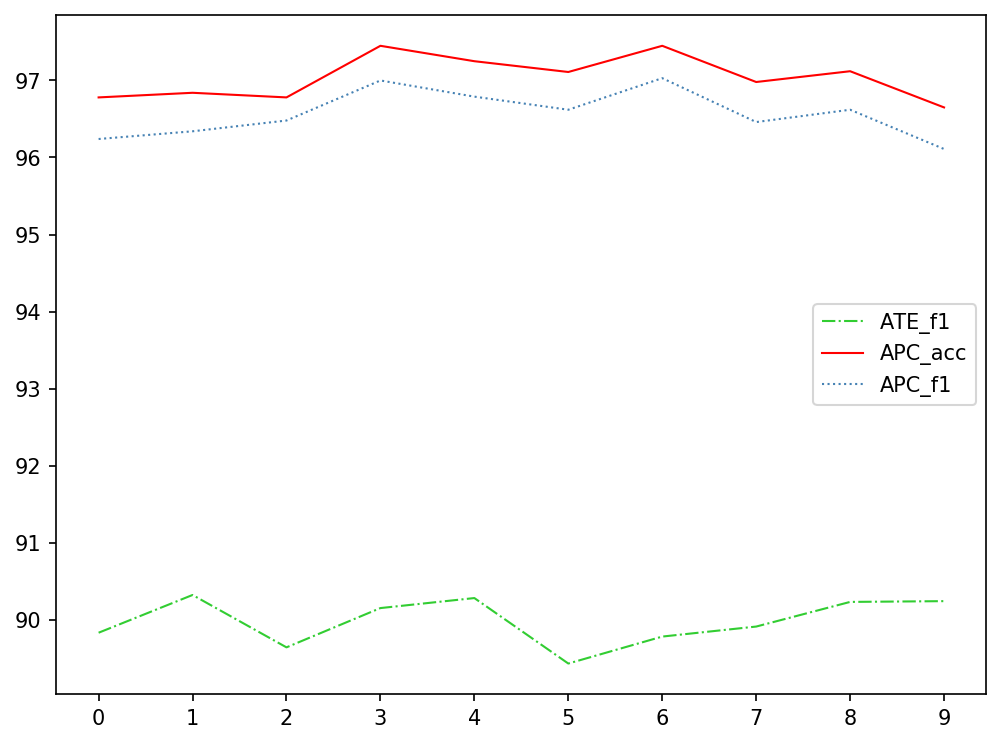,height=5cm}
		\label{fig:subfiga}
	}
	\subfigure[]
	{
		\epsfig{file=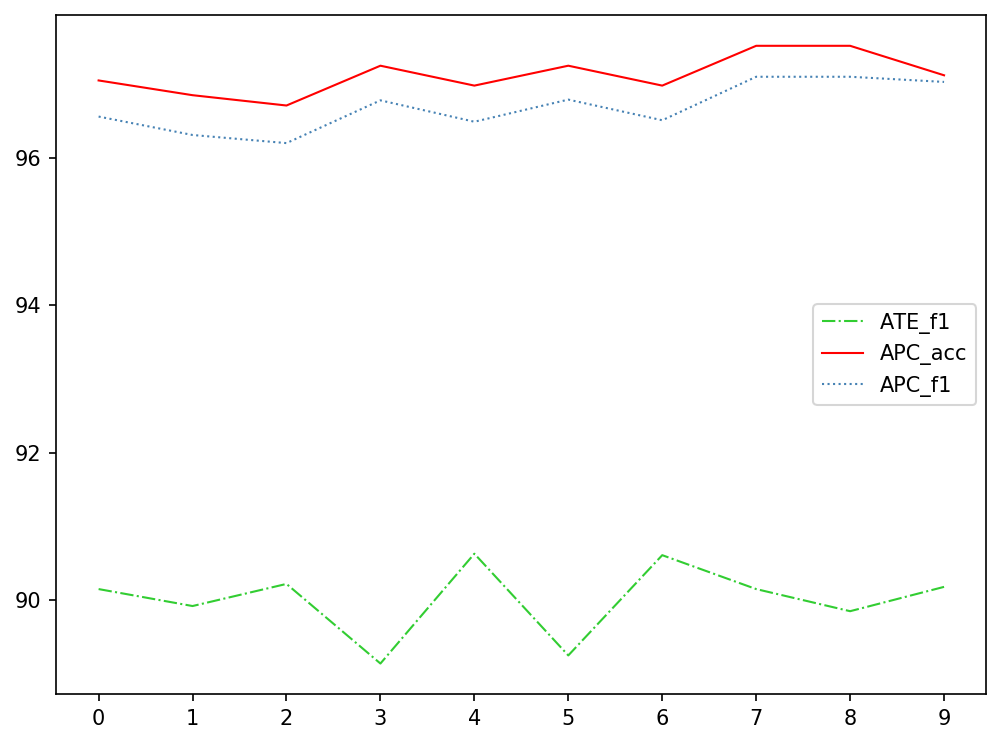,height=5cm}
		\label{fig:subfigb}
	}

	\caption{The \ref{fig:subfiga}, \ref{fig:subfigb} are the performance visualization of LCF-ATEPC-CDM and LCF-ATEPC-CDW on the Chinese Phone dataset under different SRDs, respectively.}
	\label{fig:SRDsPhone}
\end{figure*}

\begin{figure*}[pos=th]
	\centering
	\subfigure[]
	{
		\epsfig{file=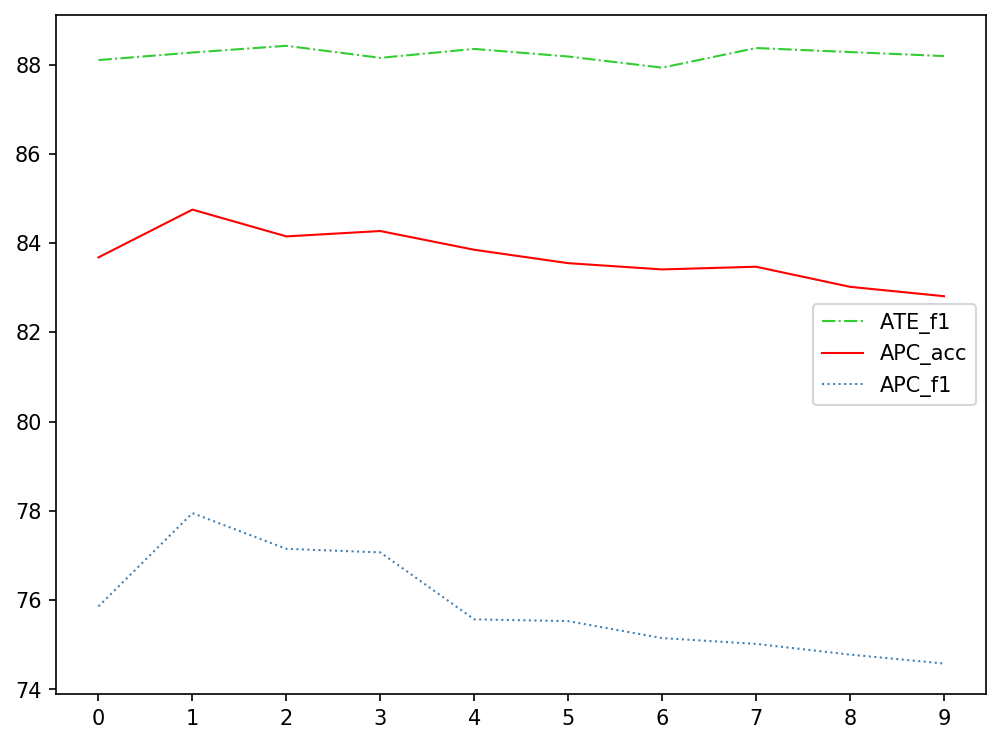,height=5cm}
		\label{fig:subfigc}
	}
	\subfigure[]
	{
		\epsfig{file=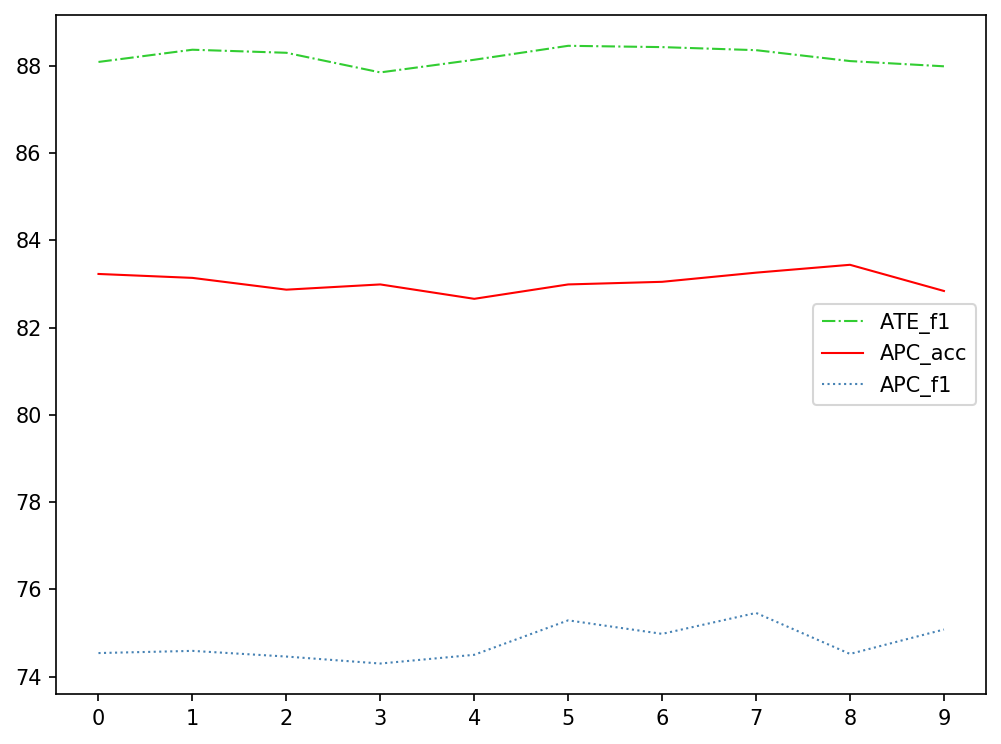,height=5cm}
		\label{fig:subfigd}
	}
	\caption{The \ref{fig:subfigc}, \ref{fig:subfigd} are the performance visualization of LCF-ATEPC-CDM and LCF-ATEPC-CDW on the Restaurant dataset under different SRDs, respectively.}
	\label{fig:SRDsRestaurant}
\end{figure*}

We tested the sensitivity of SRD threshold on the typical Chinese and English ABSA datasets: the Phone dataset and The Restaurant dataset, respectively. Besides, for the evaluation of the restaurant dataset, we adopted the domain-adapted BERT model as the underlying architecture of the LCF-ATEPC model. The experimental result of Figure \ref{fig:SRDsPhone}, \ref{fig:SRDsRestaurant} are evaluated in multi-task learning process.

For the Chinese Phone dataset, the LCF-ATEPC-CDM model can achieve the best APC accuracy and F1 score when the SRD thresholds are 3, 6, while the best ATE task performance reaches the highest when the SRD thresholds are 1, 4. The LCF-ATEPC-CDW model obtains the best APC performance on the Phone dataset when the SRD thresholds are 4 and 6, while the best ATE F1 score is approximately obtained when the SRD threshold is 7.

For the Restaurant dataset, the optimal APC accuracy and F1 score achieved by LCF-ATEPC-CDM while the SRD threshold is approximately between 2 and 4. While the SRD threshold for the LCF-ATEPC-CDW is between 5 and 7, the model achieves the optimal aspect classification accuracy and F1 score. However, the F1 score of the ATE task is less sensitive to the SRD threshold, indicating that aspect polarity classification task has less assistance on it during the joint learning process.

\section{Conclusion}
\label{sect:Conclusion}

The ATE and APC subtasks were treated as independent tasks in previous studies. Moreover, the multi-task learning model for ATE and APC subtasks has not attracted enough attention from researchers. Besides, the researches concerning the Chinese language-oriented ABSA task are not sufficient and urgent to be proposed and developed. 
To address the above problems, this paper proposes a multi-task learning model LCF-ATEPC for aspect-based sentiment analysis based on the MHSA and the LCF mechanisms and applies the pre-trained BERT to the ATE sub-tasks for the first time. 
Not only for the Chinese language, but the models proposed in this paper are multilingual and applicable to the classic English review sentiment analysis task, such as the SemEval-2014 task4.
The proposed model can automatically extract aspects from reviews and infer aspects' polarity. Empirical results on 3 commonly English datasets and four Chinese review datasets for ABSA tasks show that, compared with all models based on basic BERT, the LCF-ATEPC model achieves state-of-the-art performance on ATE and APC tasks.


\section*{Acknowledgments and Funding}
Thanks to the anonymous reviewers and the scholars who helped us. 
This research is supported by the Innovation Project of Graduate School of South China Normal University and funded by National Natural Science Foundation of China, Multi-modal Brain-Computer Interface and Its Application in Patients with Consciousness Disorder, Project approval number: 61876067. 

\bibliographystyle{cas-model2-names}
\bibliography{references}

\end{document}